\documentclass[11pt,a4paper]{article}
\usepackage[hyperref]{emnlp2020}
\usepackage{times}
\usepackage{latexsym}

\usepackage{mwe}
\usepackage{times}
\usepackage{latexsym}
\usepackage{comment}
\usepackage{amsmath}
\usepackage{url}
\usepackage{xcolor}
\usepackage{graphicx} 
\usepackage{caption}
\usepackage{subcaption}
\usepackage{booktabs,subcaption,amsfonts,dcolumn}
\usepackage{enumitem}
\usepackage{multirow,tabularx}
\newcolumntype{Y}{>{\centering\arraybackslash}X}
\renewcommand\arraystretch{2}
\newcolumntype{P}[1]{>{\centering\arraybackslash}p{#1}}
\newcolumntype{M}[1]{>{\centering\arraybackslash}m{#1}}
\usepackage{microtype}

\aclfinalcopy 
\def\aclpaperid{444444} 

\title{Understanding the Properties of Generated Corpora}

\author{Naama Zwerdling, 
Segev Shlomov, 
Esther Goldbraich, 
George Kour,
Boaz Carmeli,
\\ \bf 
Naama Tepper,
Inbal Ronen,
Vitaly Zabershinsky, 
Ateret Anaby-Tavor \\
  IBM Research \\
  \texttt{\{naamaz,Segev.Shlomov1,esthergold,naama.tepper,gkour,boazc\}@il.ibm.com},
  \\
  \texttt{\{naama.tepper,VITALYZ,atereta\}@il.ibm.com} 
}

\date{}
\begin{document}
\maketitle
\begin{abstract}
Models for text generation have become focal for many research tasks and especially for the generation of sentence corpora. However, understanding the properties of an automatically generated text corpus remains challenging.
We propose a set of tools that examine the properties of generated text corpora. Applying these tools on various generated corpora allowed us to gain new insights into the properties of the generative models. As part of our characterization process, we found remarkable differences in the corpora generated by two leading generative technologies.

\end{abstract}

\section{Introduction}

Text generation models have continued to gain traction due to groundbreaking progress in neural networks techniques \cite{serban2017hierarchical, guu2018generating, Li2016deeprl}, the recent development of advanced transformer-based architectures \cite{radford2019language}, and training over huge amounts of data \cite{wu2019conditional}.
These models all have a common \emph{generation} phase, which produces a new set of sentences based on a set of original sentences.
However, aside from traditional \cite{reiter2000building} and rule-based \cite{li2018delete} algorithms, researchers face difficulties understanding the characteristics of the generated corpora.

Researchers need to examine the generation process outcome, especially if they want to develop new generation algorithms or understand the disparity in the results when tweaking a generative model's architecture or hyper-parameters. 

In some cases, the generated corpus is used directly, for example in an education scenario that requires a diverse set of sentences to compose non-similar English quizzes. 
Alternatively, the generated sentences may serve as input for a downstream task. In this case, the best way to assess the generated corpus is by evaluating the task's end goal. In certain cases, it is beneficial to predict the quality of the generated corpus before the downstream task. For example, we might be interested in data augmentation for a text classification task in which humans label the generated sentences. The end goal here would be to improve the classifier's accuracy using the generated data. The ability to measure corpus quality and sentence readability, before applying corpus labeling could reduce the efforts required and save costs.




An automatically generated corpus could naively be evaluated using machine translation metrics such as BLEU \citeyearpar{papineni2002bleu}, ROUGE
\citeyearpar{lin2004rouge}, or METEOR \citeyearpar{lavie2007meteor}.
According to \citet{novikova2017we}, these metrics 
only weakly correlate with human ratings. In addition, these metrics only assess the quality of each single sentence but do not assess the diversity of the entire corpus or the coverage induced by the original sentences.   

To address the diversity, \citet{zhu2018texygen} introduced the notion of Self-BLEU, which compares each generated sentence to the rest of the generated set. The probabilistic approach also tackles the diversity issue. It assumes sentences are sampled from a latent distribution and uses a well-defined distribution distance measure (e.g., FID \citeyearpar{lucic2018gans} , KLD \citeyearpar{kullback1951kld}, and T-Test). 
That said, this method does not indicate whether any differences in distribution are associated with plausibility changes, mode collapse, or other issues. Mode collapse occurs when the generator generates a limited diversity of samples, or even the same sample sentence, regardless of the input. Moreover, this approach does not allow us to detect plagiarism, when the model makes small to no changes in the original sentences.

Humans are the most skillful evaluators when it comes to assessing the quality of a single sentence in terms of plausibility, fluency, grammar, relatedness to the domain, etc. However, it is much harder for humans to determine the diversity, coverage, or the distance from the original set for an entire corpus.  
Last year, \citet{hashimoto2019unifying} suggested a compelling approach that combines both human and automatic metrics. However, this approach is costly (requires 20 ratings for each sentence) and depends on the generative model itself.

In this paper we suggest a set of fully automated metrics that can help researchers understand the properties of an automatically generated corpus and assess whether it fits their task. 
Our characterization metrics cover five main dimensions:
1) Sentence quantity. 
2) Vocabulary enrichment.
3) Grammatical correctness and plausibility of the generated sentences.
4) Semantic similarity between the test and the generated datasets.
5) Syntactic distance between the original and the generated datasets.

Note that our characterization tool is unrelated to the internals of the generation model's algorithm, architecture, or parameters; it does not assume any underlying constraints on the generation process.

The tool allows us to independently assess the quality, coverage, plausibility, and diversity, in addition to evaluating the plagiarism.  
 

Using our tool and metrics, we examined several domain-specific corpora resulting from three kinds of generative models and three datasets. We compared their characteristics along the suggested metrics and discuss insights derived from the results. The code is publicly available for the NLP community.


\section{Metrics}
We address a minimal set of metrics to measure the quantity, quality, diversity, and uniqueness of a corpus of generated sentences. 
In addition to the common split of machine learning data into \textit{train}, \textit{validation}, and \textit{test}, we refer to an additional \textit{generated} set: the generation model's output sentences.

\textbf{Sentence Quantity}-- 
We measure the ability of our generation model to produce a large unique set of new sentences. 
A generated sentence is unique if no other sentence in the \textit{training} set or \textit{generated} set is identical, after lower-casing the sentence and removing punctuation.
Given a set of $n$ sentences, we denote by $U_n$ the number of unique sentences within the set. We define our quantity measure as $Unique(n)=\frac{U_n}{n} $. We can also refer to $n$ as the generation model's sampling attempts.

\newpage

\textbf{Vocabulary Augmentation}-- 
An important aspect of a generated set is the model's ability to enlarge its vocabulary beyond the one it was exposed to in the \textit{train} set. We define \textit{Vocab} as the number of new terms in the \textit{generated} set that do not exist in the \textit{train} set. The \textit{test} set's \textit{Vocab} provides a baseline for the expected \textit{Vocab} of the \textit{generated} set.

\textbf{Plausibility and Grammar}--
To estimate plausibility, we calculate the perplexity, or the likelihood of a generated sentence appearing in a language model built from a representative corpus. The representative corpus should have a large number of sentences from the same domain and style of the original set. Here, we used BERT \cite{devlin2019bert}, which was pre-trained on a large corpus of English text.
To measure the grammatical correctness of a sentence, we used a proofreading package called \textit{LanguageTool}\footnote{https://pypi.org/project/language-check/}.
The grammatical score was determined by normalizing the number of grammatical errors in the sentence over its length. 

\textbf{Semantic Similarity}--
To estimate the semantic similarity between two sets, we applied a semantic similarity measure on every pair of elements from each set and then generalized it to the set level. 
We calculated the cosine similarity between the embeddings of the two sentences to measure semantic similarity.
In this work, we chose InferSent \cite{conneau2017supervised} as our sentence embedding. 
To measure the similarity between two sets, we borrow the information retrieval notation of $Precision$ and $Recall$, similar to \citet{zhao2017learning} as follows:
\begin{align*}
Precision(G,T) &= \frac{\sum_{i=1}^{|G|}\max_{s_j\in T} Sim(s_i,s_j)}{|G|} \\ 
Recall(G,T) &=\frac{\sum_{i=1}^{|T|}\max_{s_j\in G} Sim(s_i,s_j)}{|T|} \\  
F1(G,T) &= \frac{2 \cdot (Recall \cdot Precision)}{(Recall + Precision)} 
\end{align*}
where $G$ is the \textit{generated} set of sentences and $T$ is the \textit{test} set.
The $Precision$, similar to its meaning in information retrieval, expresses the semantic relevance of each of the generated sentences to the \textit{test} set. $Recall$ expresses how well the generated set covers the \textit{test} set. A small $Recall$ may indicate mode collapse. 

\textbf{Syntactic Similarity}--
This metric measures to what extent the \textit{generated} set differs in syntax from the original \textbf{\textit{train}} set. To measure the syntactic similarity between two sentences, we use the Levenshtein word edit distance \cite{editdistance}.
Word edit distance is the number of operations (add, delete, or update term) required to transform one sentence to another. 
We will denote syntactic similarity between two sentences $s_1$ and $s_2$ as: 
\begin{equation*}
 SynSim(s_1,s_2) = \left[1+\left(\frac{EdDist(s_1,s_2)}{\max(|s_1|,|s_2|)}\right) \right]^{-1}   
\end{equation*}
where $EdDist$ is the Levenshtein word edit distance. To measure the similarity between two sets, we calculate the $Precision$, $Recall$, and $F1$ scores as is done for the semantic similarity scores. 

\section{Experiments}

We compared three different models over three corpora using our proposed metrics. We also observed the correlation between our metrics and human evaluation. We chose three corpora differing in their style, size, and domain. \textit{Amazon Reviews: Unlocked Mobile Phones}
\footnote{https://www.kaggle.com/PromptCloudHQ/amazon-reviews-unlocked-mobile-phones}
is a 40K sentence corpus with reviews of unlocked mobile phones. \textit{News Category Dataset}
\footnote{https://www.kaggle.com/rmisra/news-category-dataset}
introduced by \cite{misra2018news,misra2021sculpting} contains news headlines.
We extracted 30K political news headlines from this corpus.
\textit{Yahoo Non-Factoid Question Dataset}
\footnote{https://ciir.cs.umass.edu/downloads/nfL6/}
is a collection of questions and their corresponding answers in 21 categories. We chose 8.7K questions from the health category.

We split each corpus into train (80\%), test (10\%) and validation (10\%). We trained each model on its train set and then used the trained model to generate 100K sentences.

To examine our metrics we selected three generation models. We chose the \textbf{Easy Data Augmentation (EDA)} \citeyearpar{wei2019eda} model as a baseline. EDA is a rule-based generation model with the following actions on the training set: random term insertion, swap, and deletion in addition to synonym replacement, using Word-Net as an external data source. In addition, we trained two more neural network generation models: \textbf{Variational Autoencoder (VAE)} \citeyearpar{kingma2014auto}, an RNN autoencoder that assumes a prior distribution over a latent space. The third generation model is \textbf{GPT-2 Language Model} \citeyearpar{radford2019language}.
GPT-2 is a right-to-left language model, based on the transformer architecture, which is pre-trained with huge amounts of general language data and was later fine-tuned on our training data.

We used human evaluation to validate our results. We let 6 English speaking evaluators evaluate 900 sentences from the Amazon corpus (150 sentences from each of the train, test and 3 model generated sets). Using a 5-point Likert scale, the evaluators rated the extent to which each sentence was grammatically correct, made sense, was related to the domain, and sounds in general.

\section{Results}
The results for the Amazon corpus are shown in Table~\ref{tab:phone}. The full results are given in the Appendix. For annotation simplicity, we use the model's name for the corpus of sentences generated by the model.

\begin{table*}
\renewcommand{\arraystretch}{1.3}
\small
\centering
\begin{tabular}{||c|c|c|c|c|c|c|c|c|c|c||}
\hline
\multirow{2}{*}{Model} &\multicolumn{2}{|c|}{Quantity}&\multicolumn{2}{|c|}{Fluency}  &\multicolumn{3}{|c|}{Semantic Similarity }&\multicolumn{3}{|c|}{Syntactic Similarity}\\
\cline{2-11}
& Unique & Vocab & Grammar & Plausibility & Precision & Recall & F1 & Precision & Recall & F1\\
\hline \hline

Train   & 31650  & 9552 & 0.15 & NA & 0.78 & 0.781 & 0.781 & NA & NA & NA\\
\hline
Test & 3956 & 621 & 0.16 & 4.5 & NA & NA & NA & 0.609 & 0.609 & 0.609\\
\hline
EDA&93.2\%&3480&0.22&5.7&
0.756 &	0.75	&0.753&	0.85&	0.85&	0.85\\
\hline
VAE & 29.8\% & 0 & \textbf{0.07} & \textbf{2.6} & \textbf{0.808} & 0.764 & \textbf{0.785} & 0.648 & 0.616 & 0.632\\
\hline

GPT-2 & \textbf{98.2}\% & \textbf{17386} & 0.15 & 4.8
&0.77&\textbf{0.774}&0.772&\textbf{0.598}&\textbf{0.602}&\textbf{0.6}\\

\hline
\end{tabular}
\caption{Characterization metrics of EDA, VAE and GPT-2 models on the Amazon Reviews dataset }
\label{tab:phone}
\end{table*}

Our simple $Unique(n)$ metric clearly shows the ability of the generation models to create new unique sentences. The values in the tables show the percentile for $n=100K$. 
We also measured $Unique(n)$, for $n \in {(10K,20K,..,100K)}$ and saw that while GPT-2 and EDA managed to preserve their percentiles, the VAE percentile decreased with the size of $n$.

As expected, only those models that were pre-trained or are using an external dataset (EDA, GPT-2) were able to produce new terms, as demonstrated by the  $Vocab$ metric. Another way to gain new terms in the other models might be to use word-pieces embedding instead of word embedding.

We can see the differences in style between the corpora in the \textit{Grammar} metric. The News corpus includes news headlines with a formal style, thus the \textit{Grammar} of its train and test is lower (better) than the \textit{Grammar} of the mobile phone reviews. This was expected since the phone reviews are in a more informal style, which is characterized by typos and grammar mistakes. 
The \textit{Grammar} scores of the phone reviews and news corpora are significantly higher (unpaired t-test $p$-value$<0.01$)  for the VAE than other generated sets. Train, test and GPT-2 are significantly higher than EDA but insignificant among themselves. 
The plausibility scores are
significant between all groups and correlated nicely with the \textit{Grammar} score.

The trade-off between \textit{Precision} and \textit{Recall} is demonstrated in the \textit{Semantic Similarity} metric. VAE had significantly higher \textit{Precision} than the training set. However, it had significantly lower \textit{Recall}. This could indicate mode collapse, which is typical for VAE. GPT-2 was almost a mirror image of VAE. It managed to reach significantly higher \textit{Recall} than the other models. However, its \textit{Precision} was poor and significantly lower than VAE. EDA's \textit{Semantic Similarity} was significantly lower than that of all the others.

\begin{figure}[!h]
  \centering
  \includegraphics[width=0.8\linewidth]{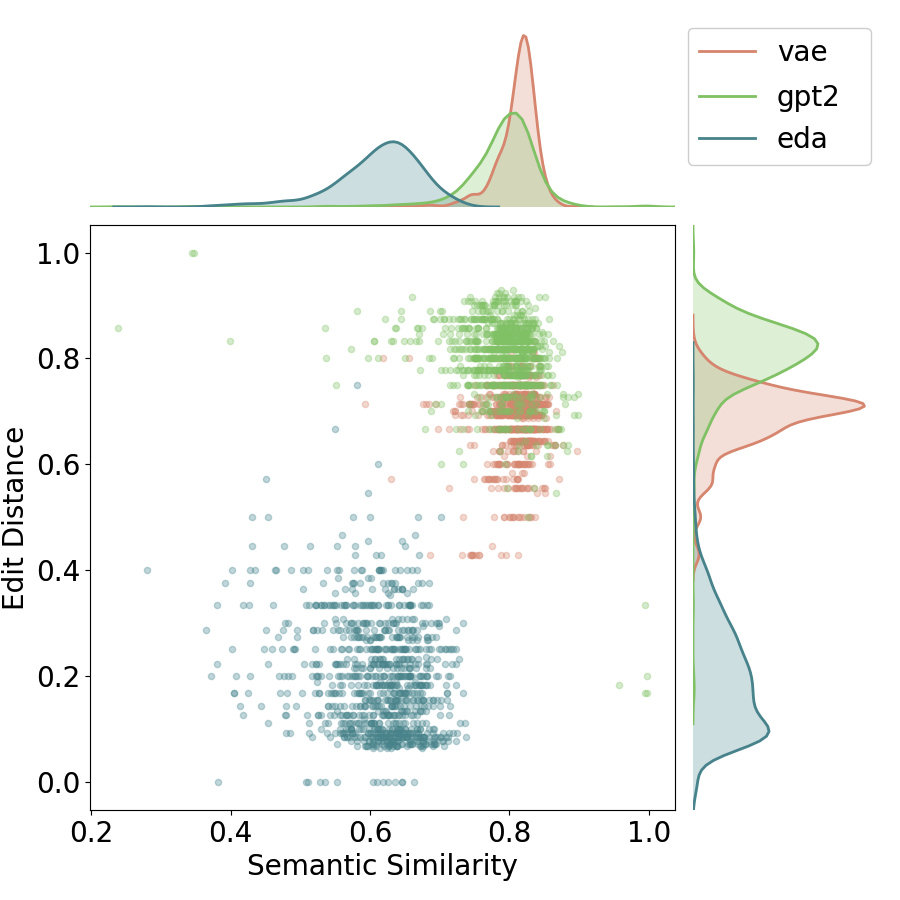}
  \caption{ News generated samples density. Each dot represents the minimal Semantic/Syntactic similarity of the sample to a sentence in the test/train set}
  \label{fig:syn_sem_phone}
\end{figure}

Our metrics also reveal a trade-off between \textit{semantic} and \textit{syntactic similarities}. As can be expected, sentences that are similar in their syntax are generally also similar in their semantics. Figure \ref{fig:syn_sem_phone} demonstrates the semantic-syntactic trade-off.

For cases where the task is text classification, one might seek out a model that produces most of its sentences in the graph's upper right corner, i.e., sentences that are semantically similar to the test and syntactically different from the train. We can also think of this graph as a way to filter out sentences by choosing those with the best trade-off.

We used human evaluators to validate our results. Notably, it is rather easy for human evaluators to assess the quality of one sentence in terms of its grammar, reasonability, and relevance to the domain. Therefore, we expected the human evaluation results to be consistent with our \textit{Grammar} and \textit{Semantic Precision} metrics. However, it is more difficult for humans to assess how well a set of sentences covers the entire domain (\textit{Recall}) or how textually close the set is to the original training set (\textit{Syntactic similarity}). Table~\ref{tab:human} includes the human evaluation results of the percentile of sentences with scores of 4 or 5.

\begin{table}[!h]
\footnotesize
\renewcommand{\arraystretch}{1.3}
\begin{tabular}{||c|c|>{\centering}m{0.4in}|>{\centering}m{0.45in}|c||}
\hline
Model & Grammar &  Make Sense  &  Domain Rel.  &  General\\
\hline \hline
Train (test) & 13 (13) & 12 (11) & 8.9 (8.3) & 9.4 (8.7)\\
\hline
EDA & 8.8 & 4.3 & 6 & 3.6\\
\hline
VAE & \textbf{14} & \textbf{11.3} & \textbf{10.3} & \textbf{9.9}\\
\hline

GPT-2 & 11.6 & 5.8 & 6.4 & 5.4\\
\hline

\end{tabular}
\caption{Human evaluation }
\label{tab:human}
\end{table}

The human evaluated grammar scores are consistent with our \textit{Grammar} metric. The \textit{Grammar}, \textit{Making Sense}, and \textit{Domain Relevance} scores show no significant difference (Mann Whitney, $p$-value $<0.01$) between train, test, and VAE. However, they are all significantly better than GPT-2. EDA has substantially inferior scores in all the above measures. 
The \textit{General} human evaluation score shows a significant correlation of 0.08 with \textit{Semantic Similarity} (Spearman, $p$-value $<0.01$). 



\section{Conclusions}

This work presented an effective set of metrics for characterizing domain-specific sentence corpora.
These metrics were able to reveal non-trivial characteristic differences between corpora and captured the divergent generation approaches of the different algorithms.
Specifically, our results show how these metrics were able to capture the tendency of the GPT-2 algorithm to generate high variability data, while the VAE was able to generate high quality but conservative data samples. The results also show how these characterization methods are well-aligned with the way humans perceive the samples in the corpora. 
Anecdotally, using these metrics, we identified a bug in one of the models while evaluating its generated corpus.


\bibliographystyle{acl_natbib}
\bibliography{anthology,emnlp2020}

\newpage
\appendix
\section{Supplemental Material}
To offer a better understanding of the metrics, we provide additional material.

\subsection{Trade-off Between Semantic and Syntactic Similarities}
\begin{figure}[!h]
\begin{subfigure}{0.99\columnwidth}
  \centering
  \includegraphics[width=0.99\linewidth]{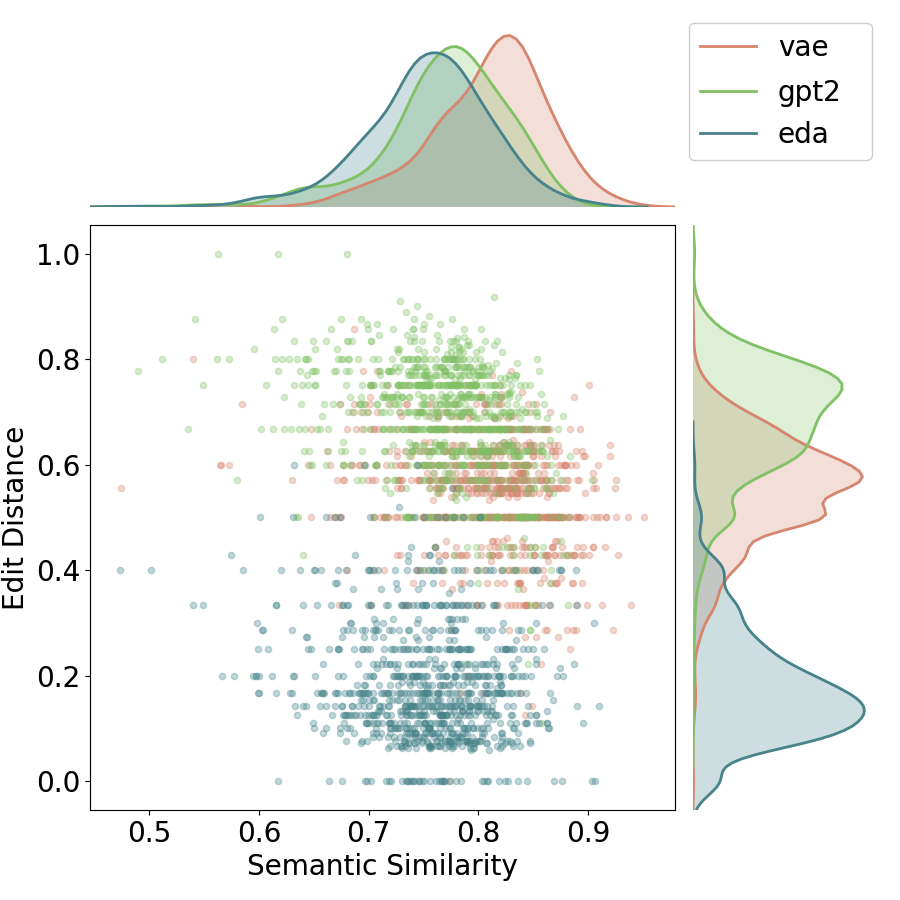}
 	\caption{Amazon}
\end{subfigure}

\begin{subfigure}{0.99\columnwidth}
  \centering
  \includegraphics[width=0.99\linewidth]{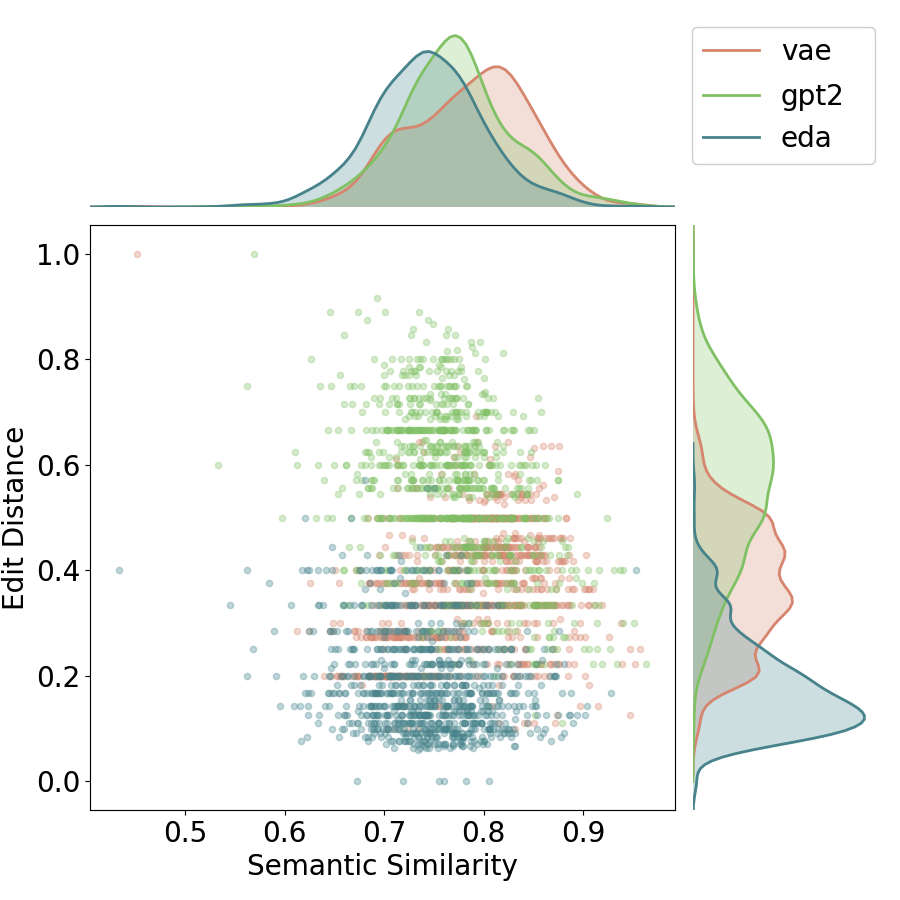}
  \caption{Yahoo! Answers}
\end{subfigure}
 \caption{Generated samples density in the semantic-syntactic plane. The horizontal axis represents the minimal semantic similarity of the sample to a sentence in the test set. The vertical axis is the normalized edit distance to the syntactically closest sentence in the train set.}
 \label{fig:syn_sem_all}
\end{figure}

Below you can find Figure \ref{fig:syn_sem_all} which is an extension to Figure \ref{fig:syn_sem_phone} with two additional datasets: Amazon News and Yahoo Answers.

\subsection{Sentences Example}
Below you can find randomly picked examples for generated sentences from each one of the three models trained on the Amazon Reviews: Unlocked Mobile Phones dataset.

\vspace{0.6cm}
\textbf{EDA}
  \begin{itemize}[noitemsep]
   \small
  \setlength\itemsep{0.5em}
    \item shipped quickly outstanding quality works just fine
    \item great phone and very only thing is camara touchsensitive
    \item works great outstanding lots of space to get apps
    \item this is not a good typesetters case it has not stay on
    \item so far so interahamwe good works well
    \end{itemize}

\vspace{0.6cm}
\textbf{VAE}
  \begin{itemize}[noitemsep]
   \small
  \setlength\itemsep{0.5em}
    \item I like this phone but I have nt had any problems with it
    \item excelente producto llego con el tiempo de I RECOMMEND
    \item great product and fast shipping great deal
    \item Did not work well Battery life is poor
    \item I like the phone Very good
    \end{itemize}

\vspace{0.6cm}
\textbf{GPT-2}
  \begin{itemize}[noitemsep]
   \small
  \setlength\itemsep{0.5em}
    \item Nice little phone .. price was right.ok
    \item Great smart phone - capable of standing up to an all day army .
   \item Phone did not work so caught the signal too much .
    \item It did my thing when the phone was advertised.works just fine now
    \item A productive phone but the battery dies really fast .
    \end{itemize}
\vspace{0.6cm}

\subsection{Characterization Metrics Table}
Below you can find Tables \ref{tab:News} and Table \ref{tab:Yahoo}, which extends Table \ref{tab:phone} with two additional datasets.

\begin{table*}
\renewcommand{\arraystretch}{1.5}
\small
\centering
\begin{tabular}{||c|c|c|c|c|c|c|c|c|c|c||}
\hline
\multirow{2}{*}{Model} &\multicolumn{2}{|c|}{Quantity}&\multicolumn{2}{|c|}{Fluency}  &\multicolumn{3}{|c|}{Semantic Similarity }&\multicolumn{3}{|c|}{Syntactic Similarity}\\
\cline{2-11}
& Unique & Vocab & Grammar & Plausibility & Precision & Recall & F1 & Precision & Recall & F1\\
\hline \hline
Train&24523&16211&0.060&NA&0.808&0.808&0.808&NA&NA&NA\\
\hline
Test&3065&850&0.058&6.36&NA&NA&NA&0.564&0.565&0.564\\
\hline
EDA&94.6\%&5405&0.258&13.688&0.612&0.606&0.609&0.842&0.842&0.842\\
\hline
VAE&6.3\%&0&\textbf{0.017}&\textbf{6.20}&\textbf{0.811}&0.777&\textbf{0.794}&0.594&\textbf{0.555}&0.574\\
\hline
GPT-2&\textbf{99.6}\%&\textbf{39104}&0.082&7.57&0.79&\textbf{0.801}&0.796&\textbf{0.556}&0.56&\textbf{0.558}\\
\hline
\end{tabular}
\caption{News }
\label{tab:News}
\vspace{1cm}
\renewcommand{\arraystretch}{1.5}
\small
\centering
\begin{tabular}{||c|c|c|c|c|c|c|c|c|c|c||}
\hline
\multirow{2}{*}{Model} &\multicolumn{2}{|c|}{Quantity}&\multicolumn{2}{|c|}{Fluency}  &\multicolumn{3}{|c|}{Semantic Similarity }&\multicolumn{3}{|c|}{Syntactic Similarity}\\
\cline{2-11}
& Unique & Vocab & Grammar & Plausibility & Precision & Recall & F1 & Precision & Recall & F1\\
\hline \hline
Train &7015&6731&0.108&NA&0.79&0.785&0.787&NA&NA&NA\\ \hline
Test &871&431&0.106&5.10&NA&NA&NA&0.662&0.656&0.659\\ \hline
EDA &84.0\%&5467&0.205&5.46&0.743&0.749&0.746&0.85&0.851&0.851\\ \hline
VAE &1.1\%&0&0.117&\textbf{1.61}&\textbf{0.787}&0.739&0.763&0.736&\textbf{0.637}&0.683\\ \hline

GPT-2 &\textbf{91.5}\%&\textbf{24816}&\textbf{0.10}3&5.13&0.771&\textbf{0.782}&\textbf{0.776}&\textbf{0.65}&0.654&\textbf{0.652}\\ \hline

\end{tabular}
\caption{Yahoo }
\label{tab:Yahoo}
\end{table*}

\newpage
t
\end{document}